\documentclass[11pt,a4paper]{article}
\usepackage{acl2019}
\usepackage{times}
\usepackage{latexsym}

\usepackage{url}
\usepackage{csquotes}

\usepackage{multirow}
\usepackage{booktabs}
\usepackage{graphicx}

\usepackage{framed}

\usepackage{yfonts}

\usepackage{amssymb} 

\aclfinalcopy 
\def\aclpaperid{21} 

\title{Towards Robust Named Entity Recognition for Historic German}

\author{Stefan Schweter and Johannes Baiter \\
Bayerische Staatsbibliothek M{\"u}nchen\\
Digital Library/Munich Digitization Center \\
80539 Munich, Germany\\
 {\tt \{stefan.schweter, johannes.baiter\}@bsb-muenchen.de}}

\date{}

\begin{document}
\maketitle
\begin{abstract}
  Recent advances in language modeling using deep neural networks have shown that these models learn representations, that vary with the network depth from morphology to semantic relationships like co-reference. We apply pre-trained language models to low-resource named entity recognition for Historic German. We show on a series of experiments that character-based pre-trained language models do not run into trouble when faced with low-resource datasets. Our pre-trained character-based language models improve upon classical CRF-based methods and previous work on Bi-LSTMs by boosting F1 score performance by up to 6\%.
  Our pre-trained language and NER models are publicly available\footnote{\url{https://github.com/stefan-it/historic-ner}}.
\end{abstract}

\section{Introduction}

Named entity recognition (NER) is a central component in natural language processing tasks. Identifying named entities is a key part in systems e.g. for question answering or entity linking. Traditionally, NER systems are built using conditional random fields (CRFs). Recent systems are using neural network architectures like bidirectional LSTM with a CRF-layer ontop and pre-trained word embeddings \citep{P16-1101,N16-1030,D17-1035,W17-4421}. 

Pre-trained word embeddings have been shown to be of great use for downstream NLP tasks ~\citep{mikolov2013distributed,pennington2014glove}. Many recently proposed approaches go beyond these pre-trained embeddings. Recent works have proposed methods that produce different representations for the same word depending on its contextual usage~\cite{P17-1161,peters2018iclr,akbik2018coling,devlin2018bert}. 
These methods have shown to be very powerful in the fields of named entity recognition, coreference resolution, part-of-speech tagging and question answering, especially in combination with classic word embeddings.

Our paper is based on the work of \citet{Riedl2018}. They showed how to build a model for German named entity recognition (NER) that performs at the state of the art for both contemporary and historical texts. Labeled historical texts for German named entity recognition are a low-resource domain. In order to achieve robust state-of-the-art results for historical texts they used transfer-learning with labeled data from other high-resource domains like CoNLL-2003 \citep{W03-0419} or GermEval \citep{L14-1251}. They showed that using Bi-LSTM with a CRF as the top layer and word embeddings outperforms CRFs with hand-coded features in a big-data situation.

We build up upon their work and use the same low-resource datasets for Historic German. Furthermore, we show how to achieve new state-of-the-art results for Historic German named entity recognition by using only unlabeled data via pre-trained language models and word embeddings. We also introduce a novel language model pre-training objective, that uses only contemporary texts for training to achieve comparable state-of-the-art results on historical texts.

\section{Model}

\begin{figure}
 \centering
 \includegraphics[width=\linewidth]{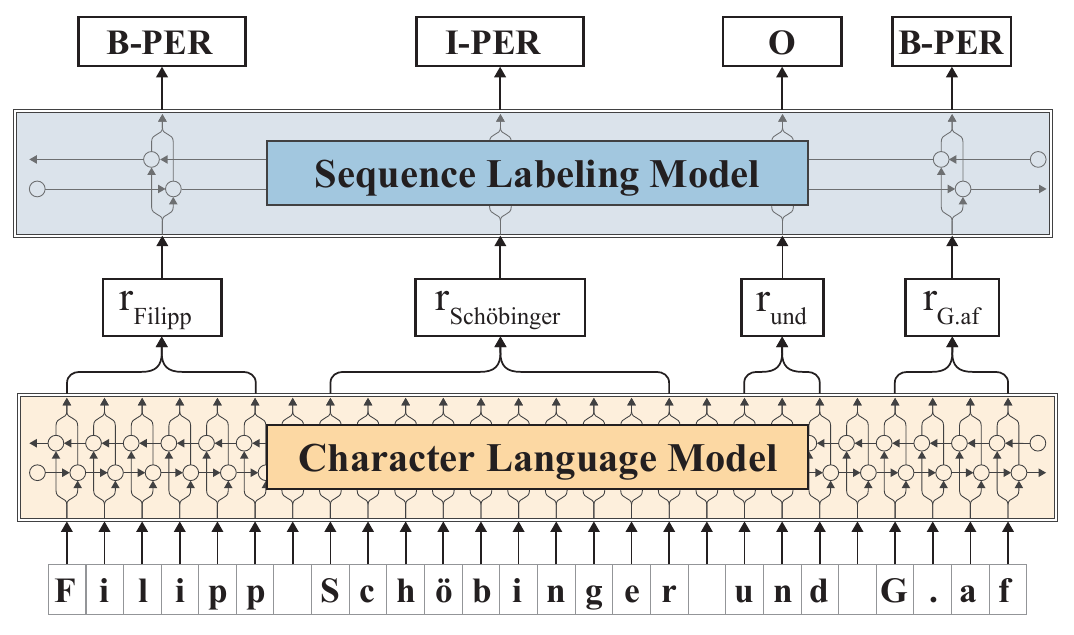}
 \captionof{figure}{High level overview of our used model. A sentence is input as a character sequence into a pre-trained bidirectional character language model. From this LM, we retrieve for each word a contextual embedding that we pass into a vanilla Bi-LSTM-CRF.}
 \label{high-level-model-overview}
\end{figure}

In this paper, we use contextualized string embeddings as proposed by \citet{akbik2018coling}, as they have shown to be very effective in named entity recognition. We use the \textsc{Flair}\footnote{\url{https://github.com/zalandoresearch/flair}} \citep{akbik2018coling} library to train all NER and pre-trained language models. We use FastText (Wikipedia and Crawl) as word embeddings. \textsc{Flair} allows us to easily combine (``stacking'') different embeddings types. For instance, \citet{lample2016neural} combine word embeddings with character features. In our experiments we combined several embedding types and language models. Contextualized string embeddings were trained with a forward and backward character-based language model (LSTM) on two historic datasets. This process is further called ``pre-training''. We use a Bi-LSTM with CRF on top as proposed by \citet{2015arXiv150801991H}. A high level system overview of our used model is shown
in figure \ref{high-level-model-overview}.

\section{Datasets}

We use the same two datasets for Historic German as used by \citet{Riedl2018}. These datasets are based on historical texts that were extracted \citep{NEUDECKER16.110} from the Europeana collection of historical newspapers\footnote{\url{https://www.europeana.eu/portal/de}}. The first corpus is the collection of Tyrolean periodicals and newspapers from the Dr Friedrich Temann Library (LFT). The LFT corpus consists of approximately 87,000 tokens from 1926. The second corpus is a collection of Austrian newspaper texts from the Austrian National Library (ONB). The ONB corpus consists of approximately 35,000 tokens from texts created between 1710 and 1873. 

The tagset includes locations (LOC), organizations (ORG), persons (PER) and the remaining entities as miscellaneous (MISC). Figures \ref{stats-onb}-\ref{stats-lft} contain an overview of the number of named entities of the two datasets. No miscellaneous entities (MISC) are found in the ONB dataset and only a few are annotated in the LFT dataset. The two corpora pose three challenging problems: they are relatively small compared to contemporary corpora like CoNLL-2003 or GermEval. They also have a different language variety (German and Austrian) and they include a high rate of OCR errors\footnote{Typical OCR errors would be segmentation and hyphenation errors or misrecognition of characters (e.g. \emph{B~i~f~m~a~r~c~k} instead of \emph{B~i} \textfrak{s} \emph{m~a~r~c~k}).} since they were
originally printed in Gothic type-face (Fraktur), a low resource font, which has not been the main focus of recent OCR research.
\begin{table}[h!]
\begin{center}
\begin{tabular}{ l c c c c }
\toprule
Dataset & LOC & MISC & ORG & PER \\
\midrule
Training & 1,605 & 0 & 182 & 2,674 \\
Development & 207 & 0 & 10 & 447 \\
Test & 221 & 0 & 16 & 355 \\
\bottomrule
\end{tabular}
\end{center}
\caption{\label{stats-onb} Number of named entities in ONB dataset.}
\end{table}

\begin{table}[h!]
\begin{center}
\begin{tabular}{ l c c c c }
\toprule
Dataset & LOC & MISC & ORG & PER \\
\midrule
Training & 3,998 & 2 & 2,293 & 4,009 \\
Development & 406 & 0 & 264 & 558 \\
Test & 441 & 1 & 324 & 506 \\
\bottomrule
\end{tabular}
\end{center}
\caption{\label{stats-lft} Number of named entities in LFT dataset.}
\end{table}

\section{Experiments}

\subsection{Experiment 1: Different Word Embeddings}

\begin{table*}[t]
\begin{center}
\begin{tabular}{ l l c }
\toprule
 & Configuration & F-Score \\

\midrule
\parbox[t]{2mm}{\multirow{6}{*}{\rotatebox[origin=c]{90}{LFT}}} & Wikipedia & 69.59\% \\
& Common Crawl & 68.97\% \\
& Wikipedia + Common Crawl &  72.00\% \\
& Wikipedia + Common Crawl + Character &  \textbf{74.50\%} \\
& \citet{Riedl2018} (no transfer-learning) & 69.62\% \\
& \citet{Riedl2018} (with transfer-learning) & 74.33\% \\

\hline

\parbox[t]{2mm}{\multirow{6}{*}{\rotatebox[origin=c]{90}{ONB}}} & Wikipedia & 75.80\% \\
& CommonCrawl & 78.70\% \\
& Wikipedia + CommonCrawl & 79.46\% \\
& Wikipedia + CommonCrawl + Character & \textbf{80.48\%} \\
& \citet{Riedl2018} (no transfer-learning) & 73.31\% \\
& \citet{Riedl2018} (with transfer-learning) & 78.56\% \\

\bottomrule
\end{tabular}
\end{center}
\caption{\label{configuration-all} Results on LFT and ONB dataset with different configurations. Wikipedia and Common Crawl are pre-trained FastText word embeddings. The best configurations reported by \citet{Riedl2018} used Wikipedia or Europeana word embeddings with subword information and character embeddings.}
\end{table*}

In the first experiment we use different types of embeddings on the two datasets: (a) FastText embeddings trained on German Wikipedia articles, (b) FastText embeddings trained on Common Crawl and (c) character embeddings, as proposed by \citet{lample2016neural}. We use pre-trained FastText embeddings\footnote{\url{https://fasttext.cc/docs/en/crawl-vectors.html}} without subword information, as we found out that subword information could harm performance (0.8 to 1.5\%) of our system in some cases.

Table \ref{configuration-all} shows, that combining pre-trained FastText for Wikipedia and Common Crawl leads to a F1 score of 72.50\% on the LFT dataset. Adding character embeddings has a positive impact of 2\% and yields 74.50\%. This result is higher than the reported
one by \citet{Riedl2018} (74.33\%), who used transfer-learning with more labeled data. Table \ref{configuration-all} also shows the same effect for ONB: combining Wikipedia and
Common Crawl embeddings leads to 79.46\% and adding character embeddings marginally improves the result to 80.48\%. This result is also higher than the reported one by \citet{Riedl2018} (78.56\%).

\subsection{Experiment 2: Language model pre-training}

\begin{table*}[h!]
\begin{center}
\begin{tabular}{l c c c c}
\toprule
& \textbf{Configuration} & Pre-trained LM & Pre-training data & \textbf{F-Score}  \\
\midrule
\parbox[t]{2mm}{\multirow{7}{*}{\rotatebox[origin=c]{90}{LFT (1926)}}} & German & \checkmark & Wikipedia, OPUS & 76.04\% \\
& Hamburger Anzeiger (HHA) & \checkmark & Newspaper (1888 - 1945) & \textbf{77.51\%} \\
& Wiener Zeitung (WZ) & \checkmark  & Newspaper (1703 - 1875) & 75.60\% \\
& Multi-lingual BERT  & \checkmark & Wikipedia & 74.39\% \\
& \textsc{smlm} (synthetic corpus) & \checkmark & Wikipedia & 77.16\%\\
& \citet{Riedl2018} & - & - & 69.62\% \\
& \citet{Riedl2018}\textsuperscript{$\dagger$} & - & - & 74.33\% \\

\hline

\parbox[t]{2mm}{\multirow{7}{*}{\rotatebox[origin=c]{90}{ONB (1710-1873)}}} & German & \checkmark & Wikipedia, OPUS & 80.06\% \\
& Hamburger Anzeiger (HHA) & \checkmark & Newspaper (1888 - 1945) & 83.28\% \\
& Wiener Zeitung (WZ) & \checkmark  & Newspaper (1703 - 1875) & \textbf{85.31\%} \\
& Multi-lingual BERT & \checkmark & Wikipedia & 77.19\% \\
& \textsc{smlm} (synthetic corpus) & \checkmark & Wikipedia & 82.15\% \\
& \citet{Riedl2018} & - & - & 73.31\% \\
& \citet{Riedl2018}\textsuperscript{$\dagger$} & - & - & 78.56\% \\
\bottomrule
\end{tabular}
\end{center}
\caption{\label{lm-all} Results on LFT and ONB with different language models. The German language model refers to the model used in \citet{akbik2018coling}. We perform a per-layer analysis for BERT on the development set and use the best layer. For all experiments we also use pre-trained FastText embeddings on Wikipedia and Common Crawl as well as character embeddings. \textsuperscript{$\dagger$} indicates the usage of additional training data (GermEval) for transfer learning.}
\end{table*}

\begin{figure}
 \centering
 \includegraphics[width=\linewidth]{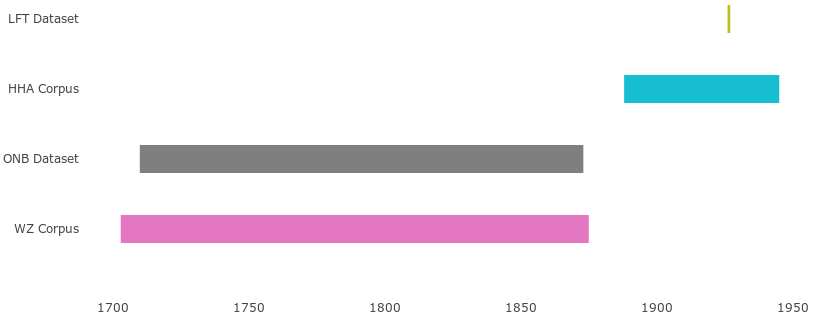}
 \captionof{figure}{Temporal overlap for language model corpora and historic datasets.}
 \label{epoch-overlap}
\end{figure}

For the next experiments we train contextualized string embeddings as proposed by \citet{akbik2018coling}. We train language models on two datasets from the Europeana collection of historical newspapers. The first corpus consists of articles from the Hamburger Anzeiger newspaper (HHA) covering 741,575,357 tokens from 1888 - 1945. The second corpus consists of articles from the Wiener Zeitung newspaper (WZ) covering  801,543,845 tokens from 1703 - 1875.
We choose the two corpora, because they have a temporal overlap with the LFT corpus (1926) and the ONB corpus (1710 - 1873). Figure \ref{epoch-overlap} shows the temporal overlap for the language model corpora and the datasets used in the downstream task. There is a huge temporal overlap between the ONB dataset and the WZ corpus, whereas the overlap between the LFT dataset and the HHA corpus is relatively small. 

Additionally we use the  BERT model, that was trained on Wikipedia for 104 languages\footnote{\url{https://github.com/google-research/bert/blob/f39e881b169b9d53bea03d2d341b31707a6c052b/multilingual.md}} for comparison. We perform a per-layer analysis of the multi-lingual BERT model on the development set to find the best layer for our task. For the German language model, we use the same pre-trained language model for German as used in \citet{akbik2018coling}. This model was trained on various sources (Wikipedia, OPUS) with a training data set size of half a billion tokens.

Table \ref{lm-all} shows that the temporal aspect of training data for the language models has deep impact on the performance. On LFT (1926) the language model trained on the HHA corpus (1888 - 1945) leads to a F1 score of 77.51\%, which is a new state-of-the art result on this dataset. The result is 3.18\% better than the result reported by \citet{Riedl2018}, which uses transfer-learning with more labeled training data. The language model trained on the WZ corpus (1703-1875) only achieves a F1 score of 75.60\%, likely because the time period of the data used for pre-training (19th century) is too far removed from that of the downstream task (mid-1920s). Table \ref{lm-all} also shows the results of pre-trained language models on the ONB (1710 - 1873) dataset. The language models, that were trained on contemporary data like the German Wikipedia \citep{akbik2018coling} or multi-lingual BERT  do not perform very well on the
ONB dataset, which covers texts from the 18-19th century. The language model trained on the HHA corpus performs better, since there is a substantially temporal overlap with the ONB corpus. The language model trained on the WZ corpus (1703-1875) leads to the best results with a F1 score of 85.31\%. This result is 6.75\% better than the reported result by \citet{Riedl2018}, which again uses transfer-learning with additionally labeled training data.

\subsection{Experiment 3: Synthetic Masked Language Modeling (\textsc{smlm})}

We also consider the masked language modeling (\textsc{mlm}) objective of \citet{devlin2018bert}. However, this technique cannot be directly used, because they use a subword-based language model, in contrast to our character-based language model. We introduce a novel masked language modeling technique, synthetic masked language modeling (\textsc{smlm}) that randomly adds noise during training.

The main motivation for using \textsc{smlm} is to transfer a corpus from one domain (e.g. \enquote{clean} contemporary texts) into another (e.g. \enquote{noisy} historical texts). \textsc{smlm} uses the vocabulary (characters) from the target domain and injects them into the source domain. With this technique it is possible to create a synthetic corpus, that ``emulates'' OCR errors or spelling mistakes without having any data from the target domain (except all possible characters as vocabulary). Furthermore, \textsc{smlm} can also be seen as a kind of domain adaption.

To use \textsc{smlm} we extract all vocabulary (characters) from the ONB and LFT datasets. We refer to these characters as target vocabulary. Then we obtained a corpus consisting of contemporary texts from Leipzig Corpora Collection \citep{Goldhahn12buildinglarge} for German. The resulting corpus has 388,961,352 tokens. During training, the following \textsc{smlm} objective is used: Iterate overall characters in the contemporary corpus.
Leave the character unchanged in 90\% of the time. For the remaining 10\% we employ the following strategy: in 20\% of the time replace the character with a masked character, that does not exist in the target vocabulary. In 80\% of the time we randomly replace the character by a symbol from the target vocabulary. 

Table \ref{lm-all} shows that the language model trained with \textsc{smlm} achieves the second best result on LFT with 77.16\%. The ONB corpus is more challenging for \textsc{smlm}, because it includes texts from a totally different time period (18-19th century). \textsc{smlm} achieves the third best result with a F-Score of 82.15\%. This result is remarkable, because the language model itself has never seen texts from the 18-19th century. The model was trained on contemporary texts with \textsc{smlm} only.

\section{Data Analysis}

\begin{table}[h!]
\begin{center}
\begin{tabular}{ l c c c c}
\toprule
 & \multirow{2}{*}{LM} & \multicolumn{2}{c}{Perplexity} & \multirow{2}{*}{F-Score}
\\
\cmidrule{3-4}
& & Forward & Backward & \\
\midrule
\parbox[t]{2mm}{\multirow{4}{*}{\rotatebox[origin=c]{90}{LFT}}} & German & 8.30 & 8.7 & 76.04\% \\
 & HHA & \textbf{6.31} & \textbf{6.64} & \textbf{77.51\%} \\
 & WZ & 6.72 & 6.97 & 75.60\% \\
 & Synthetic & 7.87 & 8.20 &  77.16\%\\

\hline

\parbox[t]{2mm}{\multirow{4}{*}{\rotatebox[origin=c]{90}{ONB}}} & German & 8.58 & 8.77 & 80.06\% \\
 & HHA & 6.71 & 7.22 & 83.28\% \\
 & WZ & \textbf{4.72} & \textbf{4.95} & \textbf{85.31\%} \\
 & Synthetic & 8.65 & 9.64 & 82.15\%\\
\bottomrule
\end{tabular}
\end{center}
\caption{\label{analysis-all} Averaged perplexity for all sentences in the test dataset for LFT for all pre-trained language models.}
\end{table}

The usage of pre-trained character-based language models boosts performance for both LFT and ONB datasets. The results in table \ref{lm-all} show, that the selection of the language model corpus plays an important role: a corpus with a large degree of temporal overlap with the downstream task performs better than corpus with little to no temporal overlap. In order to compare our trained language models with each other, we measure both the perplexity of the forward language model and the backward language model on the test dataset for LFT and ONB. The perplexity for each sentence in the test dataset is calculated and averaged. The results for LFT and ONB are shown in table \ref{analysis-all}.
For all language models (except one) there is a clear correlation between overall perplexity and F1 score on the test dataset: lower perplexity (both for forward and backward language model) yields better performance in terms of the F1 score on the downstream NER tasks. But this assumption does not hold for the language model that was trained on synthetic data via \textsc{smlm} objective: The perplexity for this language model (both forward and backward) is relatively high compared to other language models, but the F1 score results are better than some other language models with lower perplexity. This variation can be observed both on LFT and ONB test data. We leave this anomaly here as an open question: Is perplexity a good measure for comparing language models and a useful indicator for their results on downstream tasks?

The previous experiments show, that language model pre-training does work very well, even for domains with low data resources. \citet{I17-2016} showed that using CRF-based methods outperform traditional Bi-LSTM in low-resource settings. We argue that this shortcoming can now be eliminated by using Bi-LSTMs in combination with pre-trained language models. Our experiments also showed, that pre-trained language models can also help to improve performance, even when no training data for the target domain is used (\textsc{smlm} objective).




\section{Conclusion}

In this paper we have studied the influence of using language model pre-training for named entity recognition for Historic German. We achieve new state-of-the-art results using carefully chosen training data for language models.

For a low-resource domain like named entity recognition for Historic German, language model pre-training can be a strong competitor to CRF-only methods as proposed by \citet{I17-2016}. We showed that language model pre-training can be more effective than using transfer-learning with labeled datasets.

Furthermore, we introduced a new language model pre-training objective, synthetic masked language model pre-training (\textsc{smlm}), that allows a transfer from one domain (contemporary texts) to another domain (historical texts) by using only the same (character) vocabulary. Results showed that using \textsc{smlm} can achieve comparable results for Historic named entity recognition, even when they are only trained on contemporary texts.

\section*{Acknowledgments}

We would like to thank the anonymous reviewers for their helpful and valuable comments. 

\bibliography{acl2019}
\bibliographystyle{acl_natbib}

\clearpage

\appendix

\section{Supplemental Material}
\label{sec:supplemental}

\subsection{Language model pre-training}

Table \ref{lm-training-parameters} shows the parameters that we used for training our language models. As our character-based language model relies on raw text, no preprocessing steps like tokenization are needed. We use 1/500 of the complete corpus for development data and another 1/500 for test data during the language model training.


\begin{table}[h!]
\begin{center}
\begin{tabular}{ l l c }
\toprule
Parameter & Value \\
\midrule
LSTM hidden size & 2048 \\
LSTM layer & 1 \\
Dropout & 0.1 \\
Sequence length (characters) & 250 \\
Mini batch size & 1 \\
Epochs & 1 \\
\bottomrule
\end{tabular}
\end{center}
\caption{\label{lm-training-parameters} Parameters used for language model pre-training.}
\end{table}

\subsubsection{\textsc{smlm} objective}

\begin{figure}[h!]
\begin{framed}
\textbf{Original sentence}

\small{Dann habe der Mann erz\"ahlt, wie er in M\"unchen am Bahnhof mit Blumen begr\"u\ss{}t worden sei.}

\textbf{Sentence after \textsc{smlm} transformation}

\small{
Qa \P n hab5 der MaRy erz\"ahlt nie er in M\"nchenIam Bahnhof mit Blumen begr\"u\ss{}( Corden se\P.
}

\end{framed}
\caption{An example of the \textsc{smlm} transformation for a given input sentence. The special character ``\P'' is used as masked character symbol.}
\label{smlm-example}
\end{figure}

Figure \ref{smlm-example} shows the \textsc{smlm} objective for a given input sentence and the corresponding output. 
We use the same parameters as shown in table \ref{lm-training-parameters} to train a language model with \textsc{smlm} objective. We use different values of $p$ in range of $[80, 90, 95]$ for leaving the character unchanged in the \textsc{smlm} objective and found that $p = 90$ yields the best results.

\subsection{Model parameters}

Table \ref{ner-training-parameters} shows the parameters that we use for training a named entity recognition model with the  \textsc{Flair} library. We reduce the learning rate by a factor of 0.5 with a patience of 3. This factor determines the number of epochs with no improvement after which learning rate will be reduced.

\begin{table}[h!]
\begin{center}
\begin{tabular}{ l l c }
\toprule
Parameter & Value \\
\midrule
LSTM hidden size & 512 \\
Learning rate & 0.1 \\
Mini batch size & 8 \\
Max epochs & 500 \\
Optimizer & SGD \\
\bottomrule
\end{tabular}
\end{center}
\caption{\label{ner-training-parameters} Parameters used for training NER models.}
\end{table}

\subsection{BERT per-layer analysis}

\begin{figure}
 \centering
 \includegraphics[width=\linewidth]{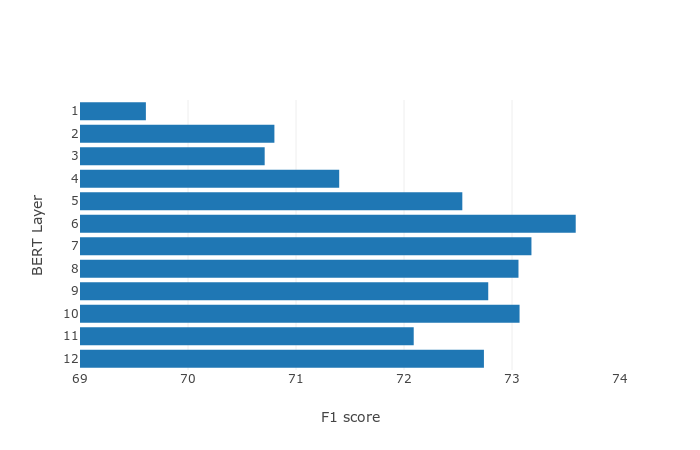}
 \captionof{figure}{BERT per-layer analysis on the LFT development dataset.}
 \label{bert-lft}
\end{figure}

\begin{figure}
 \centering
 \includegraphics[width=\linewidth]{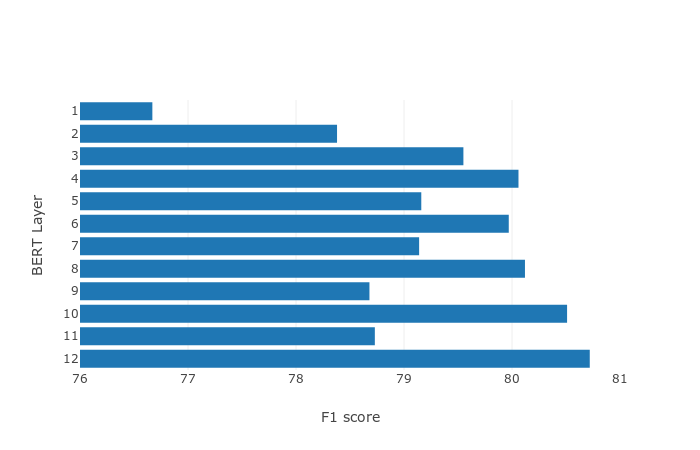}
 \captionof{figure}{BERT per-layer analysis on the ONB development dataset.}
 \label{bert-onb}
\end{figure}

We experimentally found that using the last four layers as proposed in \citet{devlin2018bert} for the feature-based approach does not work well. Thus, we perform a per-layer analysis that trains a model with a specific layer from the multi-lingual BERT model. Inspired by \citet{D18-1179} we visualize the performance for each layer of the BERT model. Figure \ref{bert-lft} shows the performance of each layer for the LFT development dataset, figure \ref{bert-onb} for the ONB development dataset.

\subsection{Evaluation}

We train all NER models with IOBES \citep{W09-1119} tagging scheme. In the prediction step we convert IOBES tagging scheme to IOB, in order to use the offical CoNLL-2003 evaluation script\footnote{\url{https://www.clips.uantwerpen.be/conll2003/ner/bin/conlleval}}. For all NER models we train and evaluate 3 runs and report an averaged F1 score.

\subsection{Negative Results}

We briefly describe a few ideas we implemented that did not seem to be effective in initial experiments. These findings are from early initial experiments. We did not pursue these experiments further after first attempts, but some approaches could be effective with proper hyperparameter tunings.

\begin{itemize}
    \item \textbf{FastText embeddings with subword information}: We use subword information with FastText embeddings trained on Wikipedia articles. On LFT this model was 0.81\% behind a model trained with FastText embeddings without subword information. On ONB the difference was 1.56\%. Using both FastText embeddings trained on Wikipedia and CommonCrawl with subword information caused out-of-memory errors on our system with 32GB of RAM.
    \item \textbf{ELMo Transformer}: We trained ELMo Transformer models as proposed by \citet{D18-1179} for both HH and WZ corpus. We use the default hyperparameters as proposed by \citet{D18-1179} and trained a ELMo Transformer model for one epoch (one iteration over the whole corpus) with a vocabulary size of 1,094,628 tokens both for the HH and WZ corpus. We use the same model architecture like in previous experiments for training a NER model on both LFT and ONB. On LFT we achieved a F1 score of 72.18\%, which is 5.33\% behind our new state-of-the-art result. On ONB we achieved a F1 score of 75.72\%, which is 9.59\% behind our new state-of-the-art result. We assume that training a ELMo Transformer model for more epochs would lead to better results.
\end{itemize}

\end{document}